\documentclass{article}
\usepackage{ijcai25}

\usepackage{times}
\usepackage{soul}
\usepackage{url}
\usepackage[hidelinks]{hyperref}
\usepackage[utf8]{inputenc}
\usepackage[small]{caption}
\usepackage{graphicx}
\usepackage{amsmath}
\usepackage{amsthm}
\usepackage{booktabs}
\usepackage{algorithm}
\usepackage[switch]{lineno}

\usepackage{natbib}
\usepackage{amssymb}
\usepackage{subcaption}
\usepackage{multirow}
\usepackage{algpseudocode}
\usepackage{xcolor}

\newcommand{\tr}{\mathrm{tr}}
\newcommand{\avgacc}{\overline{\mathrm{acc}}}
\newcommand{\avgtft}{\mathrm{tft}}
\newcommand{\avgtrt}{\mathrm{trt}}

\newtheorem{definition}{Definition}
\newtheorem{result}{Result}

\algrenewcommand\algorithmicrequire{\textbf{Input:}}
\algrenewcommand\algorithmicensure{\textbf{Output:}}

\title{Sequence Transferability and Task Order Selection in Continual Learning}

\author{
\vspace{5mm}
\qquad Thinh Nguyen$^1$\footnote{Corresponding author.} \and
Cuong N. Nguyen$^2$ \and
Quang Pham$^3$\footnote{The author contributed to this work while working at A*STAR.} \and
Binh T. Nguyen$^4$ \newline
Savitha Ramasamy$^5$\and
Xiaoli Li$^5$ \and
Cuong V. Nguyen$^2$ \\[5mm]
\affiliations
$^1$VinUniversity, Vietnam\\
$^2$Durham University, United Kingdom\\
$^3$Independent Researcher\\
$^4$VNU-HCM, University of Science, Vietnam\\
$^5$Institute for Infocomm Research, A*STAR, Singapore \\[5mm]
\emails
\textit{thinh.nth@vinuni.edu.vn}
}

\begin{document}

\maketitle

\begin{abstract}
In continual learning, understanding the properties of task sequences and their relationships to model performance is important for developing advanced algorithms with better accuracy. However, efforts in this direction remain underdeveloped despite encouraging progress in methodology development. In this work, we investigate the impacts of sequence transferability on continual learning and propose two novel measures that capture the total transferability of a task sequence, either in the forward or backward direction. Based on the empirical properties of these measures, we then develop a new method for the task order selection problem in continual learning. Our method can be shown to offer a better performance than the conventional strategy of random task selection.
\end{abstract}

\section{Introduction}
\label{sec:intro}

Continual learning (CL)~\citep{ring1994continual} has emerged as an important and challenging research problem toward artificial general intelligence~\citep{parisi2019continual}. Aiming at building systems that can interact with the environment to accumulate knowledge over time, CL offers solutions to real-world applications~\citep{diethe2019continual}, opening up several promising research directions~\citep{hayes2021replay}. In recent years, extensive efforts have been devoted to developing state-of-the-art CL strategies~\citep{de2021continual, masana2022class, carta2023comprehensive} and their theoretical properties~\citep{pentina2015lifelong, knoblauch2020optimal, lin2023theory}. Despite the encouraging progress, practical CL solutions still require a deep understanding of the relationships between characteristics of the given task sequence and the final model performance, which remains underdeveloped.

Along this direction, \citet{nguyen2019toward} pioneered a study demonstrating that the total complexity of a task sequence correlates with the average accuracy of CL algorithms. Their findings suggest that sequence complexity can serve as a proxy for CL performance, providing valuable insights for designing benchmarks and algorithms in continual learning. However, their work considered each task independently, disregarding their interrelationships. While this assumption simplifies their analyses, it neglects the sequential nature of tasks, which is crucial for CL. Recent theoretical studies \citep{prado2022theory, lin2023theory} attempted to derive error bounds and the expected forgetting for CL using different notions of task similarity. However, these similarity metrics, such as the $\mathcal{F}$-relatedness in \citet{prado2022theory} or the Euclidean distance in \citet{lin2023theory}, are impractical to utilize in large-scale real-world applications. Therefore, it is important to develop a principled yet practical approach to quantify the sequential information among CL tasks and analyze how it affects the model performance.

In this work, we generalize the notion of transferability measures (or transferability metrics) in transfer learning \citep{nguyen2020leep, you2021logme, huang2022frustratingly}, which have been successful in efficiently predicting the effectiveness of knowledge transfer between two tasks, to develop novel transferability measures for sequences of tasks in CL.
In particular, we propose two sequence transferability measures, \emph{Total Forward Transferability} (TFT) and \emph{Total Reverse Transferability} (TRT), that employ a conventional transferability metric to quantify the forward transfer and forgetting of CL algorithms in a data-dependent manner. Inheriting the advantages of conventional transferability metrics, our methods are easy to compute and can be shown empirically to be well-correlated with average accuracies of CL algorithms.


These properties of our new measures reveal intriguing relationships between sequence transferability and CL performance, which we leverage to develop a novel algorithm for the continual task order selection problem. This problem arises from real-world scenarios where data can be accumulated over a period of time to form a batch of tasks~\citep{caccia2022anytime} before being fed to a CL algorithm. In this case, training CL models in the chronological task order may not be optimal and thus it is desirable to develop methods for selecting better task orders. In our paper, we use the idea of the total forward transferability to develop the \emph{Heuristic Continual Task Order Selection} (HCTOS) method for solving this problem. Our method can be demonstrated to select good task orders for CL while being robust to hyper-parameter choices such as the sample size or its base transferability metric.


In summary, our work makes the following contributions: (1) we propose TFT and TRT, two novel sequence transferability measures that have close associations with CL performance; and (2) we develop the HCTOS algorithm to address the continual task order selection problem in practical CL that can yield better learning accuracy while being robust to hyper-parameter configurations.

\section{Related Works}

Our study is relevant to several machine learning fields, including continual learning, transfer learning, and especially transferability estimation. In the following, we provide an overview of previous works directly related to ours.

\subsection{Continual Learning}

CL aims at building and continuously adapting a model through a sequence of tasks so that the model can perform well on all tasks it has observed~\citep{ring1994continual,thrun1995lifelong}. However, when the model is updated on new tasks, SGD training, which assumes i.i.d.~data, usually leads to a decline in performance on earlier tasks. This phenomenon is referred to as catastrophic forgetting~\citep{mccloskey1989catastrophic} and has been the primary subject of research on CL. Numerous strategies have been suggested to mitigate this issue, including regularization-based methods~\citep{farajtabar2020orthogonal, yin2020optimization, bian2024make}, episodic memory-based methods~\citep{chaudhry2019tiny, buzzega2020dark, bellitto2024saliency}, parameter isolation-based methods~\citep{yoon2017lifelong, mallya2018packnet, tang2025mind}, and Bayesian methods~\citep{nguyen2024lifelong, hai2024continual}. Besides algorithmic efforts to reduce catastrophic forgetting, some recent works also look at the role of forward transfer~\citep{vinyals2016matching, snell2017prototypical} and backward transfer~\citep{lin2022beyond, wan2022continual}.

Although a learner may retain information from previous tasks, it is often more crucial to employ the acquired knowledge to efficiently learn new tasks (referred to as forward transfer) and to retain the previous knowledge (known as backward transfer)~\citep{lopez2017gradient}. \citet{raghavan2021formalizing} argued that the enhancement in accuracy does not necessarily improve the forward transfer and backward transfer. On the other hand, \cite{prado2022theory} showed that the relatedness between tasks can affect the model's performance. In this paper, we will explore the relationships between accuracy, forward transfer, and backward transfer in CL. To the best of our knowledge, the most similar to our work is \cite{chen2023forgetting}, which also showed that less forgetting leads to better forward transfer, but their models are overconfident on the trained auxiliary output layer and are more time-consuming to train.

In terms of task selection, real-world data can be accumulated to form a batch of tasks~\citep{caccia2022anytime}, and training them sequentially in chronological order may not be optimal. Therefore, optimizing the model's performance requires selecting the right tasks in the right order. \cite{bell2022effect} indicated that task orders could significantly impact CL algorithms' effectiveness. In this work, we also develop a novel algorithm to determine a task order that can give a good accuracy for CL algorithms.

\subsection{Transferability Measures}

CL can be considered a generalization of transfer learning where we need to continuously transfer knowledge from previous tasks to current ones. In transfer learning, transferability measures~\citep{nguyen2020leep, you2021logme, yang2023pick} have emerged as a tool to estimate the easiness of knowledge transfer from a source to a target task. These measures could be used to analyze task relations~\citep{achille2019task2vec}, ranking checkpoints~\citep{li2021ranking}, or model selection~\citep{you2021logme}. Recent attempts in this area~\citep{tran2019transferability, nguyen2020leep, you2021logme, li2023exploring} aim to develop transferability measures that are well-correlated with the test accuracy of the target task. 

Label-based estimators~\citep{tran2019transferability, nguyen2020leep} rely on the relationship between source and target labels to construct transferability measures. However, they could be restrictive due to the assumption that source and target tasks shared the same inputs~\citep{tran2019transferability} or could be inaccurate due to the overfitting of the last layer to source domains~\citep{nguyen2020leep}. These issues are addressed in subsequent works using the feature layer. For instance, LogME~\citep{you2021logme} measures the transferability based on the log marginal likelihood of the target labels given the extracted features. TransRate~\citep{huang2022frustratingly} uses the mutual information between extracted features and labels to estimate transferability. Other works compute the domain distance~\citep{tan2021otce}, intra-class distance~\citep{xu2023fast}, or combine with inter-class distance~\citep{bao2019information, pandy2022transferability} to construct their measures. Recently, physics-inspired and energy-based methods~\citep{gholami2023etran, li2023exploring} have emerged as a promising approach to this problem. In this paper, we shall generalize this notion of transferability measure to define the sequence transferability measures, which can estimate the transferability of CL algorithms on a sequence of tasks.

\section{Sequence Transferability Measures and Their Associations with CL Performance}
\label{sec:correlation}

In this section, we aim to develop measures for quantifying the characteristics of a task sequence such that they are well-correlated with the average accuracy of CL algorithms. A straightforward approach is to use the accuracy difference before and after training~\citep{lopez2017gradient}. However, since this metric can only be calculated post-training, it may not be practical in applications such as model selection or task order selection~\citep{rajasegaran2019random,bell2022effect}. Instead, we shall develop measures that are based on the transferability metrics in transfer learning \citep{nguyen2020leep, you2021logme, huang2022frustratingly}. Our measures utilize these existing metrics as a base component to determine the transferability of task sequences in CL.

\subsection{Settings and Notations}
\label{subsec:notation}

Assume we have a sequence of $T$ tasks for task incremental learning, where each task $t \in \{ 1, 2, \ldots, T \}$ has a corresponding true but unknown data distribution $\mathbb{P}_t (X_t, Y_t)$. Here $X_t$ and $Y_t$ are random variables for the input and label of the $t^\mathrm{th}$ task respectively. Following the usual setting for CL in practice, we assume all $X_t$'s share the same range but each $Y_t$ may have their own range, i.e., $X_t \in \mathcal{X}$ and $Y_t \in \mathcal{Y}_t$ for all $t$.

In the task incremental learning setting, we are given a sequence of training datasets $D = \left(D_1, D_2, \ldots, D_T\right)$, where ${ D_t = \{ (x^{(i)}_t, y^{(i)}_t) \}_{i=1}^{N_t} }$ and $(x^{(i)}_t, y^{(i)}_t) \stackrel{\mathrm{iid}}{\sim} \mathbb{P}_t$. That is, for each task $t = 1, 2, \ldots, T$, the training dataset $D_t$ contains $N_t$ labeled examples drawn i.i.d. from $\mathbb{P}_t (X_t, Y_t)$. In this setting, a CL algorithm is given each dataset $D_1, D_2, \ldots, D_T$ sequentially and after learning from $D_t$, the algorithm needs to return a model $m_t$ capable of making predictions on all tasks $i \le t$. Due to the requirements for CL algorithms to avoid catastrophic forgetting while minimizing training time at each step, the model $m_t$ is usually an update of the previous model $m_{t-1}$ using some transfer learning techniques. We can also think of $m_t$ as the state of a single (possibly multi-head) CL model after observing task $t$. In this paper, we define a CL algorithm $\mathcal{A}$ as a function mapping a sequence of training datasets $D = \left(D_1, D_2, \ldots, D_T\right)$ to the sequence of models returned at each step:
\begin{equation*}
    \mathcal{A} (D) = (m_1, m_2, \ldots, m_T).
\end{equation*}

To evaluate CL algorithms, a commonly used metric is the average accuracy of the final model on all tasks that it has observed~\citep{carta2023comprehensive, wang2023comprehensive}. Formally, the \emph{average accuracy} of a CL algorithm $\mathcal{A}$ is defined as the average of the expected accuracies of $m_T$ over all tasks:

\begin{equation}
\label{eq:avgacc}
\begin{aligned}
    \avgacc(\mathcal{A}) &= \frac{1}{T} \sum_{t=1}^T \mathbb{E}_{(x,y) \sim \mathbb{P}_t} [ \mathbf{1}(y = m_T (x)) ] \\
    &= \frac{1}{T} \sum_{t=1}^T \mathbb{P}_t [ y = m_T (x) ],
\end{aligned}
\end{equation}
where $\mathbf{1}(\cdot)$ is the indicator function. In practice, we often estimate the expected accuracy $\mathbb{E}_{(x,y) \sim \mathbb{P}_t} [ \mathbf{1}(y = m_T (x)) ]$ empirically using a held-out dataset not observed during training.

\subsection{Sequence Transferability Measures}
\label{subsec:seq-trans}

In this section, we propose two new measures of sequence transferability for CL algorithms. These measures are defined based on conventional transferability measures between two tasks. Our measures are useful since they can be shown empirically to have good positive correlations with the average accuracy of CL algorithms.

We recall that a transferability measure between a pre-trained source model and a target task is a computationally inexpensive real-valued function that tells us the efficacy of transfer learning from the source model to the target task \citep{you2021logme, huang2022frustratingly, nguyen2023simple}. Adapting the definition by \citet{nguyen2023simple} to the classification setting, we can define a transferability measure as follows.

Assume we are given a pre-trained source model $m$ and a target task specified by a target training set $D_{\text{target}}$ and a true target data distribution $\mathbb{P}_{\text{target}}(X, Y)$. Let $m_{\text{target}}$ be the model after transferring $m$ to the target task by running transfer learning with $D_{\text{target}}$. A transferability measure between the source model $m$ and the target task is a real-valued function $\tr(m, D_{\text{target}}) \in \mathbb{R}$ such that $\tr(m, D_{\text{target}}) \le \tr(m', D'_{\text{target}})$ if and only if $\mathbb{P}_{\text{target}} (y = m_{\text{target}} (x)) \le \mathbb{P}'_{\text{target}} (y = m'_{\text{target}} (x))$. In other words, we can use a transferability measure as a proxy for test accuracy to compare different source models or different target tasks. In practice, the above condition rarely holds and existing transferability measures only try to approximate this ideal condition as much as possible, often via improving the correlations between $\tr(m, D_{\text{target}})$ and $\mathbb{P}_{\text{target}} (y = m_{\text{target}} (x))$.

Using a transferability measure $\tr(m, D_{\text{target}})$ above as a base transferability metric, we can now define our measures of transferability over a sequence of tasks for CL. The first measure that we propose is called the \textbf{Total Forward Transferability (TFT)} and is defined as follows.

\begin{definition}
Let $D = \left(D_1, D_2, \ldots, D_T\right)$ be a sequence of $T$ training datasets and $\mathcal{A}$ be a CL algorithm such that ${ \mathcal{A}(D) = (m_1, m_2, \ldots, m_T) }$. The total forward transferability of $\mathcal{A}$ over $D$ is:
\begin{equation}
\avgtft (\mathcal{A}, D) = \frac{1}{T-1} \sum_{t=2}^T \tr (m_{t-1}, D_t).
\label{eq:tft}
\end{equation}
\end{definition}

From the definition, TFT measures how easy it is to learn a new task for a CL algorithm. This easiness is measured by the average forward transferability, the transferability of the continual model to the next task in the sequence. Intuitively, whenever a new task arrives, TFT captures the transferability from the current model to this task. Therefore,  
we expect the TFT to capture the model's capabilities in leveraging past knowledge to facilitate the learning of new tasks throughout the CL process.

In addition to the forward transferability, it is also of interest to consider the backward transferability of the final model to a previous task. This notion of backward transferability is particularly useful because of the CL evaluation procedure: at the end of training, the final model is evaluated on all previous tasks. Thus, a high backward transferability may indicate that the model can use new information to improve its past knowledge.
To quantify this, we propose another sequence transferability measure called the \textbf{Total Reverse Transferability (TRT)}, which is defined below.

\begin{definition}
Let $D = \left(D_1, D_2, \ldots, D_T\right)$ be a sequence of $T$ training datasets and $\mathcal{A}$ be a CL algorithm such that ${ \mathcal{A}(D) = (m_1, m_2, \ldots, m_T) }$. The total reverse transferability of $\mathcal{A}$ over $D$ is:
\begin{equation}
\avgtrt (\mathcal{A}, D) = \frac{1}{T-1} \sum_{t=1}^{T-1} \tr (m_T, D_t).
\end{equation}
\end{definition}

By definition, TRT averages transferability of the final model $m_T$ to all previous tasks. In essence, it measures a CL algorithm's ability to quickly adapt to previous tasks after training, which can be considered as a measure of forgetting.
Together with TFT, they capture the degree of forgetting and forward knowledge transfer, which characterizes the stability-plasticity dilemma in CL.

In our definitions of TFT and TRT, the base transferability metric $\tr(\ldots)$ can be chosen from any existing transferability metrics such as LEEP~\citep{nguyen2020leep}, LogME~\citep{you2021logme}, TransRate~\citep{huang2022frustratingly}, etc. In this work, we will mainly use LogME due to its quality and stable performance, especially for data source selection in transfer learning scenarios~\citep{agostinelli2022stable}.

\subsection{Associations Between TFT and TRT with Performance of CL Algorithms}
\label{sec:associations}

To evaluate the effectiveness of TFT and TRT as sequence transferability measures, we conduct experiments using several CL algorithms and datasets to investigate the associations between TFT and TRT with the average accuracy of CL algorithms. The details of our experiments are given in Section~\ref{sec:seqtrans-exp}. Here we give a summary of our empirical findings.

\begin{result}
\label{obs:compare_algo}
CL algorithms with higher average accuracy generally have higher TFT and TRT values.
\end{result}

\begin{result}
\label{obs:compare_seq}
Given a CL algorithm, its average accuracy on different random task sequences is well-correlated with the corresponding TFT and TRT values.
\end{result}

These two results confirm the usefulness of TFT and TRT as proxies for comparing the effectiveness of different CL algorithms on a task sequence or comparing the hardness of different task sequences for a CL algorithm. But it is worth noting that to compute TFT and TRT, we have to actually run a CL algorithm. This is in contrast with the traditional transferability measures, which we do not need to run an actual transfer learning algorithm to compute. Developing sequence transferability measures that do not require CL training is hard in general and is a subject of future research. Nevertheless, the TFT and TRT values can still be approximated by using a simpler model and a small subset of the training data. Once this simple model is trained, calculating TFT and TRT only involves model inference, which is still computationally efficient. In this work, we mainly use TRT as a reference measure due to its similarity to the CL evaluation process. On the other hand, we will use the idea of TFT to develop a novel algorithm for continual task order selection, a challenging and important problem when applying CL in the real world.

\section{Continual Task Order Selection}
\label{sec:select}

In practical applications of CL, especially in large organizations, tasks are rarely fed to CL algorithms in a fixed order. Instead, they are often collected over different episodes and then given to a CL algorithm in batches. For instance, all the tasks that arrive each day could be collected and then given to the CL algorithm at the end of the day without any meaningful order. If the exact arriving time is available for each task, we may use it to determine an order for the tasks. However, such an order may not be optimal and the arriving time may not even be available in many cases.

To deal with this practical problem, we propose to consider the \emph{continual task order selection} (CTOS) problem, which is formalized as follows. Consider a possibly infinite sequence of batches $B_1, B_2, \ldots$, where each batch $B_i$ contains $n_i$ \emph{unordered} tasks $\{ T_{i,t} \}_{t=1}^{n_i}$. Given a CL algorithm $\mathcal{A}$, for each $i \in \{ 1, 2, \ldots \}$, find an ordered task sequence ${ S_i = (T_{i,t_1}, T_{i,t_2}, \ldots, T_{i, t_{n_i}}) }$ that maximizes $\avgacc(\mathcal{A}, S_i)$, where $(t_1, t_2, \ldots, t_{n_i})$ is a permutation of $(1, 2, \ldots, n_i)$ and $\avgacc(\mathcal{A}, S_i)$ is the average accuracy of $\mathcal{A}$ on the task sequence $S_i$.

Solving the above CTOS problem is hard in general since a brute force search requires us to compare $(n_i)!$ task permutations for each batch $B_i$. Thus, to develop a tractable solution for this problem, we propose to consider a heuristic approach that is inspired by the idea of TFT and insights from previous research in the literature.

\subsection{Single-Batch Setting}

To have a glimpse of our approach, let us consider the simple setting where there is only one batch $B$ that contains $n$ tasks, and each task $T_t$ has a corresponding training set $D_t$, with $t \in \{ 1, 2, \ldots, n \}$. For any ordered task sequence $S = (T_{t_1}, T_{t_2}, \ldots, T_{t_n})$, if we run the CL algorithm $\mathcal{A}$ on the corresponding sequence of training sets ${ D_S = (D_{t_1}, D_{t_2}, \ldots, D_{t_n}) }$, we can obtain a sequence of models $\mathcal{A}(D_S) = \{ m_{t_1}, m_{t_2}, \ldots, m_{t_n} \}$ with the TFT determined by Eq.~\eqref{eq:tft}:
\begin{equation}
\avgtft (\mathcal{A}, D_S) = \frac{1}{n-1} \sum_{i=2}^n \tr (m_{t_{i-1}}, D_{t_i}).
\label{eq:1-batch-tft}
\end{equation}

From our results in Section~\ref{sec:associations}, to find a task sequence with a high average accuracy, we can find a sequence with a high TFT score. According to Eq.~\eqref{eq:1-batch-tft}, we can construct this sequence heuristically by trying to maximize $\tr (m_{t_{i-1}}, D_{t_i})$ at every iteration $i$. From the CL and transferability estimation literature, it has been observed empirically that: (a) transferring from a hard task to an easy task often leads to a high transferability score \citep{tran2019transferability}; and (b) learning harder tasks first may enhance knowledge transfer to future tasks in CL \citep{mahdisoltani2018more}. Thus, a sensible approach to find a good task sequence is to select, at every iteration $i$, the task $T_{t_i}$ that \textbf{minimizes} the total transferability scores to all remaining tasks in the batch. By doing so, after training on this task, the future transferability scores $\tr (m_{t_{j-1}}, D_{t_j})$, with $j > i$, will likely be high, leading to a high TFT score for the whole sequence.

\begin{algorithm}[t]
\caption{Heuristic continual task order selection in \\ one batch}
\label{algorithm:1batch}
\begin{algorithmic}[1]
\Require An unordered batch of $n$ tasks $B = \{ T_t \}_{t=1}^n$.
\Statex \hspace{12pt} Training set $D_t$ for each task $T_t$.
\Statex \hspace{12pt} Continual learning algorithm $\mathcal{A}$.
\Ensure An ordered sequence $S$ of all $n$ input tasks.
\vspace{5pt}
\For{$t = 1 \textbf{ to } n$}
    \State Train a simple model $m_t$ using a subset of $D_t$.
    \State Compute $a_{t, i} := \tr (m_t, D_i)$ for all $i \ne t$. 
\EndFor
\vspace{5pt}
\State $S := [ \, ]$ ~ \text{(the empty sequence)}
\For{$i = 1 \textbf{ to } n$}
    \State Compute $L_t := \sum_{j \notin S \cup \{ t \}} a_{t, j}$, for all task $t \notin S$.
    \State Choose $t_i := \arg\min_{t \notin S} L_t$.
    \State $S := S + [ T_{t_i} ]$ ~ (concatenate $T_{t_i}$ to $S$).
\EndFor
\State \Return $S$.
\end{algorithmic}
\end{algorithm}

The procedure for implementing this idea is given in Algorithm~\ref{algorithm:1batch}. Firstly, we use a simple model and a base transferability metric to compute the transferability scores between all pairs of different tasks in the batch (lines 1--4). Then for each iteration $i = 1, 2, \ldots, n$, we choose the task index $t_i$ that minimizes the total transferability scores from task $T_{t_i}$ to all the other unselected tasks in the batch (lines 5--10).

\subsection{Multiple-Batch Setting}

In the general CTOS problem, there is a sequence of batches for which we need to select the task orders. It is straightforward to extend the procedure in Algorithm~\ref{algorithm:1batch} to this general setting. In particular, given a sequence of batches $B_1, B_2, \ldots$, we can sequentially apply Algorithm~\ref{algorithm:1batch} to each batch to select its task order. This approach gives a heuristic solution for the general CTOS problem, and thus we call it the \emph{Heuristic Continual Task Order Selection} (HCTOS) method.

In principle, the performance of HCTOS could depend on how we train the simple models and how we choose the base transferability metric in Algorithm~\ref{algorithm:1batch}. However, in our experiments in Section~\ref{sec:hctos-exp}, we can show empirically that HCTOS is generally robust to these choices.

\section{Experiments}
\label{sec:experiments}

In this section, we shall conduct experiments to evaluate TFT and TRT as sequence transferability measures. We shall also investigate the effectiveness of HCTOS for the continual task order selection problem.

\subsection{Evaluation of TFT and TRT}
\label{sec:seqtrans-exp}

In the following experiments, we investigate the effectiveness of TFT and TRT as sequence transferability measures. Specifically, we conduct two experiments to study the associations between the average accuracy of CL algorithms and their TFT or TRT values. We first describe the benchmarks and CL algorithms used in the experiments.

\subsubsection{Benchmarks}

We use the following three datasets that are widely adopted in previous CL work~\citep{zenke2017continual, lopez2017gradient, rebuffi2017icarl, wu2019large, chaudhry2019tiny, buzzega2020dark, boschini2022class}.

\paragraph{$\bullet$ Split mutual-CIFAR-10.}  The original CIFAR-10 dataset \citep{krizhevsky2009learning} contains 50,000 training images in 10 classes, with 5,000 images per class. To explore the insights in a complicated setting, we split this training set into 5 subsets corresponding to 5 tasks. Each task has three classes, with one mutual class between any two consecutive tasks.

\paragraph{$\bullet$ Split CIFAR-100.} This is a popular CL benchmark constructed by dividing the original CIFAR-100 dataset, which contains 50,000 training images from 100 classes, into 10 disjoint subsets corresponding to 10 tasks. In this way, each task has 5,000 training images from 10 distinct classes, and each class has 500 images.

\paragraph{$\bullet$ Split tiny-ImageNet.} This is another popular CL benchmark based on the Tiny-ImageNet dataset, which is a subset of ImageNet that contains 100,000 training images of 200 real objects. We follow the settings in~\citet{buzzega2020dark} to form the Split tiny-ImageNet CL benchmark, where the original dataset is split into ten non-overlapping subsets. We consider each subset as a task whose images are labeled by 20 different classes and each class has 500 samples.

\subsubsection{CL Algorithms}

We conduct our experiments on the following common replay-based CL strategies, which have been a stable approach to CL with promising performances.

\noindent $\bullet$ \textbf{Averaged GEM (A-GEM)} \citep{chaudhry2018efficient}. A-GEM is an efficient variant of Gradient Episodic Memory (GEM)~\citep{lopez2017gradient}, which uses an episodic memory to store past tasks' data and ensures that updates do not increase the loss on these tasks. A-GEM improves over GEM by supporting positive backward transfer without the high costs associated with GEM.

\noindent $\bullet$ \textbf{Experience Replay (ER)} \citep{chaudhry2019tiny}. ER uses an episodic memory to mitigate catastrophic forgetting in CL by repetitively training on examples from past tasks alongside new data, which improves generalization ability.

\noindent $\bullet$ \textbf{Dark Experience Replay ++ (DER++)} \citep{buzzega2020dark}. DER++ combines rehearsal with knowledge distillation and regularization. It stores the network's logits from past experiences in a replay buffer, helping maintain consistency across previously learned tasks and reducing forgetting, even under a limited resource scenario.

\noindent $\bullet$ \textbf{eXtended Dark Experience Replay (X-DER)}~\citep{boschini2022class}. X-DER enhances DER by adding novel insights about past data and preparing for future unseen classes. This approach leads to improved accuracy by updating past memorized experiences and adapting the model to predict future tasks, using a strategy that mimics human memory dynamics.

Other non-replay strategies, such as LwF \citep{li2017learning} and SI \citep{zenke2017continual}, have limited knowledge transfer capabilities since they do not have a mechanism to store previous data. Thus, we believe they cannot accurately reflect the transferability of task sequences. Nevertheless, for completeness, we provide their results in the appendix.

\begin{figure*}[t]
    \centering
    \includegraphics[width=.4\textwidth]{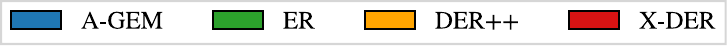}\\
    \vspace{0.2cm}
     \begin{subfigure}[b]{0.31\textwidth}
         \centering
         \includegraphics[width=\textwidth]{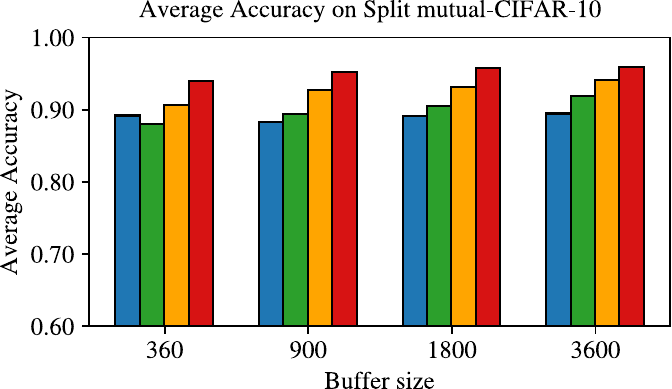}\\
         \vspace{0.3 cm}
         \includegraphics[width=\textwidth]{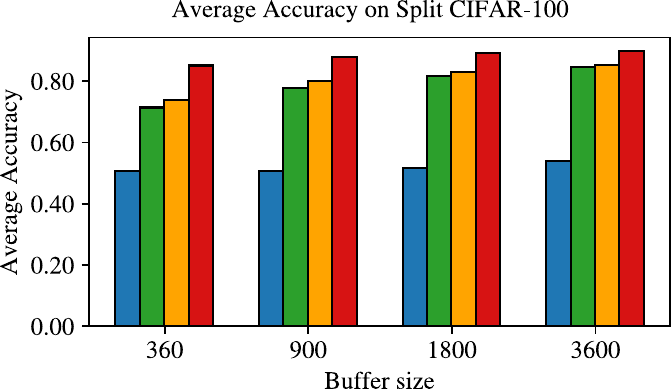}\\
         \vspace{0.3 cm}
         \includegraphics[width=\textwidth]{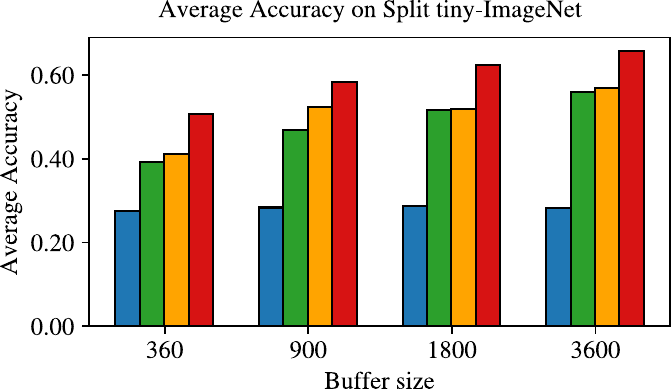}\\
         \vspace{0.2 cm}
         \caption{AA ($\uparrow$ better)}
         \label{subfig:acc:all}
     \end{subfigure}
     \hfill
     \begin{subfigure}[b]{0.31\textwidth}
         \centering
         \includegraphics[width=\textwidth]{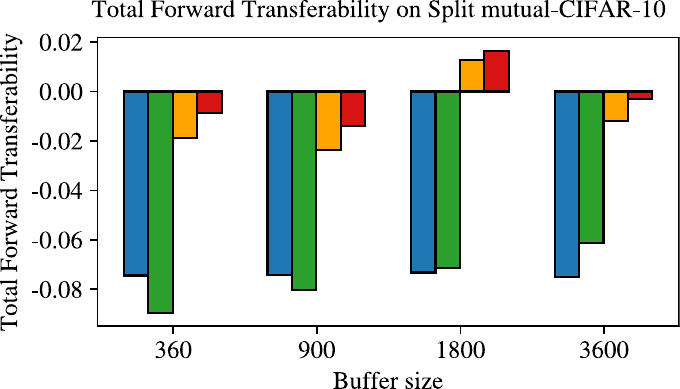}\\
         \vspace{0.3 cm}
         \includegraphics[width=\textwidth]{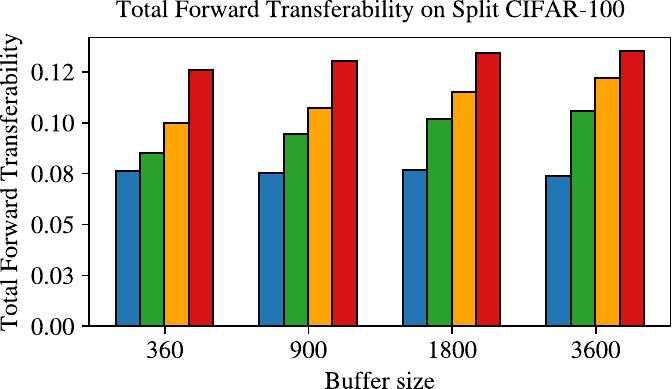}\\
         \vspace{0.3 cm}
         \includegraphics[width=\textwidth]{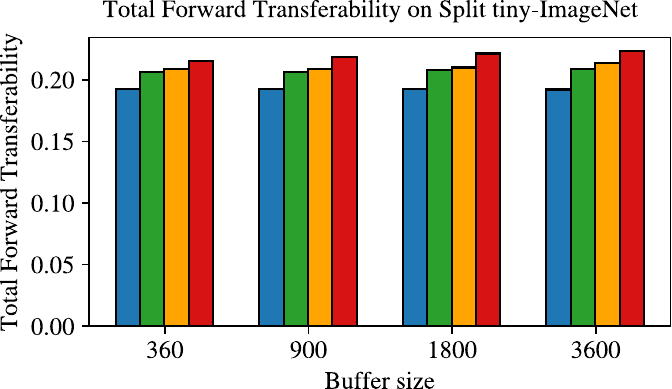}\\
         \vspace{0.2 cm}
         \caption{TFT ($\uparrow$ better)}
         \label{subfig:flogme:all}
     \end{subfigure}
     \hfill
     \begin{subfigure}[b]{0.31\textwidth}
         \centering
         \includegraphics[width=\textwidth]{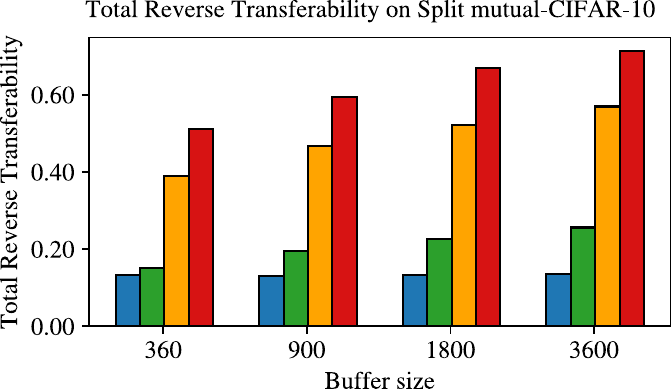}\\
         \vspace{0.3 cm}
         \includegraphics[width=\textwidth]{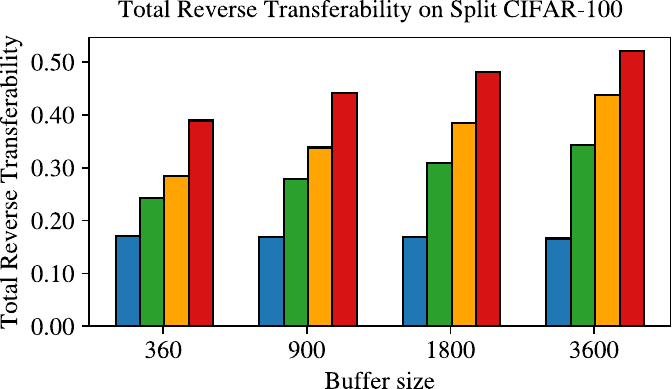}\\
         \vspace{0.3 cm}
         \includegraphics[width=\textwidth]{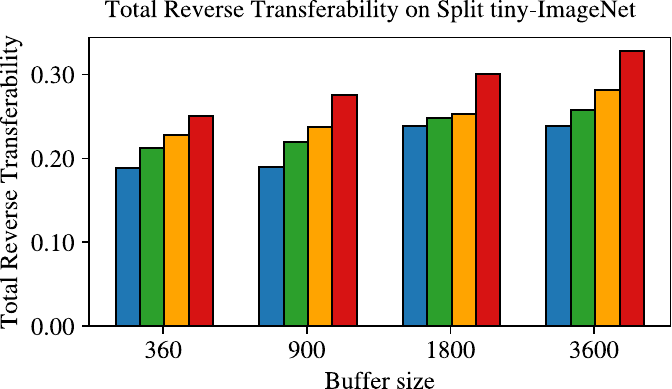}\\
         \vspace{0.2 cm}
         \label{subfig:rlogme:all}
         \caption{TRT ($\uparrow$ better)}
     \end{subfigure}
        \caption{The AA, TFT, and TRT of different replay-based CL algorithms over three benchmarks. Given a buffer size, TFT and TRT show the same trend as AA.}
        \label{fig:score:all}
        \vspace{-2mm}
\end{figure*}

\subsubsection{Experiment Setup}

In all experiments, we use ResNet18~\citep{he2016deep} trained using the SGD optimizer~\citep{bottou1998online} with a learning rate of 0.03. We train the model for 50 epochs per task on all datasets except for tiny-ImageNet, where we run 100 epochs. For fairness, we set the number of original data and replay data on their batches to 32 for all algorithms. We also adopt the hyper-parameter configurations in \citet{buzzega2020dark}. Since the performance of replay-based methods strongly depends on the buffer size, we consider four buffer sizes of 360, 900, 1800, and 3600 total samples to examine the robustness of our results. In our tables and figures, we report the average results over 20 different randomly sampled sequences, where the tasks are ordered randomly. This random ordering is essential for the generalizability of our empirical findings.

We use LogME~\citep{you2021logme} as the base transferability metric to compute TFT and TRT. To assess the associations between these sequence transferability measures and the average accuracy, we also employ the Pearson's correlation coefficient as in \citet{nguyen2019toward} and subsequent work to quantitatively estimate how strongly these quantities are correlated. A Pearson's coefficient close to 1 (or -1) indicates a strong positive (or negative) linear correlation between two quantities. On the other hand, a coefficient close to 0 suggests a weak correlation.

\subsubsection{Results}
\label{subsec:obs}

\begin{table*}[ht]
\centering
\small
\begin{tabular}{clllllllll}
\toprule
\multirow{2}{*}{Dataset} & \multirow{2}{*}{Method} & \multicolumn{4}{c}{AA and TFT} & \multicolumn{4}{c}{AA and TRT} \\ 
\cmidrule(lr){3-6} \cmidrule(lr){7-10}
& & 360 & 900 & 1800 & 3600 & 360 & 900 & 1800 & 3600 \\ 
\midrule
\multirow{4}{*}{Split mutual-CIFAR-10} & A-GEM & $0.66^{*}$ & $0.52^{*}$ & $0.71^{**}$ & $0.72^{**}$ & $0.82^{**}$ & $0.61^{*}$ & $0.93^{**}$ & $0.82^{**}$ \\
& ER & $0.47$ & $0.49$ & $0.56^{*}$ & $0.83^{**}$ & $0.65^{*}$ & $0.58^{*}$ & $0.66^{*}$ & $0.68^{*}$ \\
& DER++ & $0.45^{\dagger}$ & $0.36^{\dagger}$ & $0.57^{*}$ & $0.66^{*}$ & $0.60^{*}$ & $0.56^{*}$ & $0.68^{*}$ & $0.68^{*}$ \\
& X-DER & $0.71^{**}$ & $0.80^{**}$ & $0.76^{**}$ & $0.83^{**}$ & $0.91^{**}$ & $0.90^{**}$ &  $0.87^{**}$ &  $0.82^{**}$ \\ 
\midrule
\multirow{4}{*}{Split CIFAR-100} & A-GEM & $0.64^{*}$ & $0.59^{*}$ & $0.76^{**}$ & $0.49^{\dagger}$ & $0.87^{**}$ & $0.67^{*}$ & $0.73^{**}$ & $0.54^{\dagger}$ \\
& ER & $0.60^{*}$ & $0.54^{\dagger}$ & $0.82^{**}$ & $0.82^{**}$ & $0.67^{*}$ & $0.41^{\dagger}$ & $0.93^{**}$ & $0.87^{**}$ \\
& DER++ & $0.60^{*}$ & $0.79^{**}$ & $0.54^{*}$ & $0.45$ & $0.71^{**}$ & $0.76^{**}$ & $0.64^{*}$ & $0.65^{*}$ \\
& X-DER & $0.93^{**}$ & $0.60^{\dagger}$ & $0.90^{**}$ & $0.89^{**}$ & $0.98^{**}$ & $0.60^{\dagger}$ & $0.90^{**}$ & $0.71^{**}$ \\
\midrule
\multirow{4}{*}{Split tiny-ImageNet} & A-GEM & $0.45^{\dagger}$ & $0.40^{\dagger}$ & $0.43^{\dagger}$ & $0.66^{*}$ & $0.53^{\dagger}$ & $0.53^{\dagger}$ & $0.66^{*}$ & $0.85^{**}$ \\
& ER & $0.69^{*}$ & $0.38^{\dagger}$ & $0.46^{\dagger}$ & $0.68^{*}$ & $0.69^{*}$ & $0.46$ & $0.79^{**}$ & $0.86^{**}$ \\
& DER++ & $0.44$ & $0.37^{\dagger}$ & $0.41^{\dagger}$ & $0.53^{*}$ & $0.50^{*}$ & $0.45^{\dagger}$ & $0.57^{\dagger}$ & $0.70^{**}$ \\
& X-DER & $0.49$ &  $0.37^{\dagger}$ & $0.45^{\dagger}$ & $0.54^{*}$ & $0.74^{**}$ &  $0.62^{*}$ & $0.46^{\dagger}$ & $0.56^{*}$ \\ 
\bottomrule
\end{tabular}
\caption{Pearson's correlation coefficients between AA and TFT and between AA and TRT of four different replay-based CL algorithms with various buffer sizes. Double asterisks (**) indicate high correlations with coefficients at least 0.7. Single asterisks (*) indicate moderate correlations with coefficients between 0.5 and 0.7. Daggers ($\dagger$) indicate non-statistically significant values with $p \geq 0.05$.}
\label{table:correlation}
\vspace{-2mm}
\end{table*}

We first investigate the associations between TFT, TRT, and average accuracy (AA) for different CL algorithms (Result~\ref{obs:compare_algo} in Section~\ref{sec:associations}). In this experiment, we compare the trends of TFT, TRT, and AA of the four CL algorithms with different buffer sizes on the three CL benchmarks. From Figure~\ref{fig:score:all}, we can observe the following trend: \emph{given a buffer size, CL algorithms with higher AA have higher TFT and TRT values}. Moreover, this trend holds for all the datasets and buffer sizes, except only for the TFT with buffer size 900 on the Split mutual-CIFAR-10 where the ER algorithm has a lower TFT than A-GEM. This observation confirms our Result~\ref{obs:compare_algo} in Section~\ref{sec:associations}. As a consequence, the TFT and TRT can be used to compare different CL algorithms. In Figure~\ref{fig:score:all}, we also notice that the TFT values on Split mutual-CIFAR-10 are negative. This is because we use LogME as the base transferability metric that can return negative values~\citep{you2021logme}.

We next conduct an experiment to confirm Result~\ref{obs:compare_seq} in Section~\ref{sec:associations}. Specifically, for each CL algorithm with a given buffer size, we compare the Pearson's correlation coefficients between AA and TFT as well as between AA and TRT. These coefficients are computed over 20 random task sequences for each benchmark. The result for this experiment is given in Table~\ref{table:correlation}, which shows that \emph{AA has good correlations with TFT and TRT in the majority of settings}. Particularly, the correlation coefficients are in the moderate to very high range in more than 70\% of the settings, with more than half of them in the high range (i.e., coefficient values at least 0.7). This supports our Result~\ref{obs:compare_seq} in Section~\ref{sec:associations}. As a result, the TFT and TRT can be used to compare the hardness of different task sequences given a CL algorithm with a specified buffer size. In Table~\ref{table:correlation}, we also observe that the correlations between AA and TRT are typically greater than those between AA and TFT. One possible explanation is that both AA and TRT are computed using the final model $m_T$ returned by a CL algorithm, which makes these two quantities more correlated.

\subsection{Evaluation of HCTOS}
\label{sec:hctos-exp}

In this set of experiments, we evaluate the effectiveness of our HCTOS approach for the CTOS problem. We consider the Split CIFAR-100 benchmark and conduct our experiments using DER++ with a buffer size of 360. For the simple models in Algorithm~\ref{algorithm:1batch}, we use ResNet18 as the model architecture and restrict to only 20 random samples per class for training. We also use LogME as the base transferability metric.

\begin{figure}[t]
    \centering
    \includegraphics[height=3.25cm]{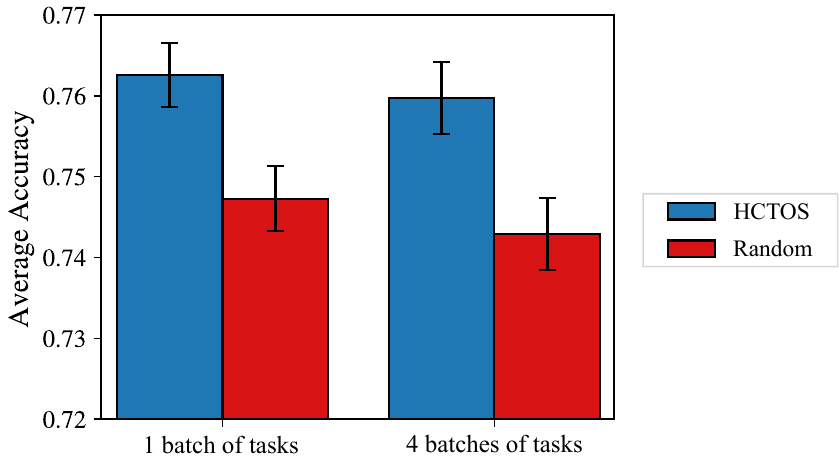}
    \vspace{-1mm}
    \caption{Comparison of AA on task sequences selected by HCTOS and random strategies. Our HCTOS method shows better performance in both single-batch and multiple-batch settings.}
    \label{fig:task_selection}
    \vspace{-1mm}
\end{figure}

We consider two settings for the experiments. The first setting has only one batch containing 20 tasks while the second setting has four batches with 5 tasks per batch. We compare HCTOS with the random selection, where the task order for each batch is chosen randomly. In real-world scenarios, this baseline could represent, e.g., the strategy where tasks are ordered based on their receiving timestamps. Figure~\ref{fig:task_selection} shows the results for these experiments, where we observe that our algorithm can select better task sequences than the random baseline in both single-batch and multiple-batch settings.

\begin{figure}[t]
    \centering
    \includegraphics[height=3.2cm]{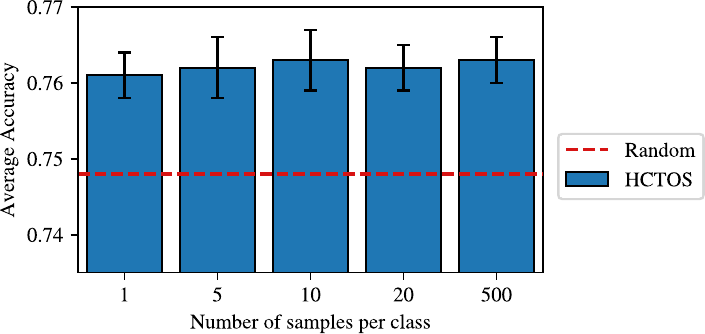}
    \vspace{-1mm}
    \caption{Effects of sample size when training simple models on the performance of HCTOS. Our method is robust to the sample size and is consistently better than the random baseline.}
    \label{fig:task_selection_sample}
    \vspace{-1mm}
\end{figure}

We also conduct two additional experiments to investigate the robustness of HCTOS to the sample size when calculating the transferability scores and to the choice of the base transferability metric. In the first experiment, we vary the number of samples per class used to train the simple models in our algorithm and examine its effects on AA. In Figure~\ref{fig:task_selection_sample}, we give the result for this experiment where our algorithm is shown to be highly robust to the sample size. Notably, even when using only one sample per class to train the simple models, our algorithm only suffers a slight decay in accuracy. The figure also shows that our algorithm is better than the random baseline for all sample sizes.

\begin{figure}[t]
    \centering
    \includegraphics[scale=0.47]{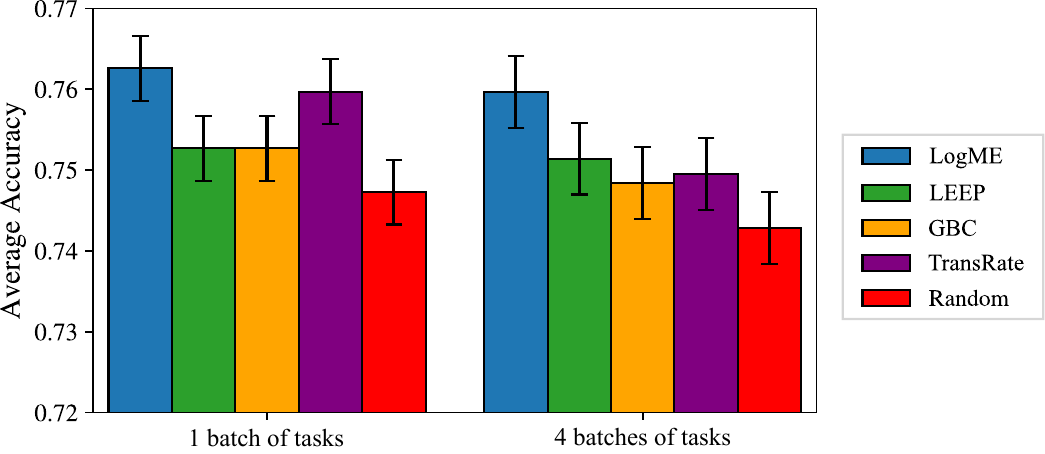}
    \vspace{-1mm}
    \caption{Average accuracy of HCTOS with respect to four different base transferability metrics. LogME exhibits higher performance than its counterparts.}
    \label{fig:task_selection_transfer}
    \vspace{-1mm}
\end{figure}

\begin{figure*}[ht]
    \centering
    \includegraphics[scale=0.65]{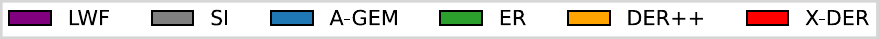}\\
    \vspace{0.2cm}
     \begin{subfigure}[b]{0.31\textwidth}
         \centering
         \includegraphics[width=\textwidth]{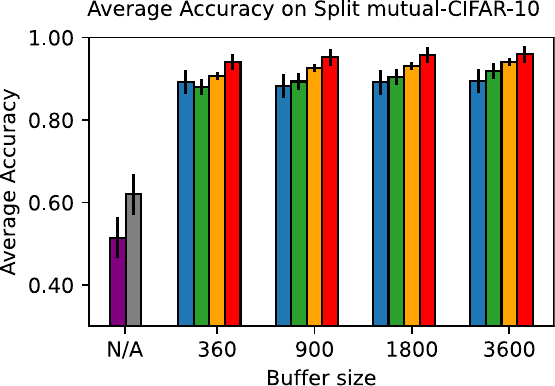} \\
         \vspace{0.3 cm}
         \includegraphics[width=\textwidth]{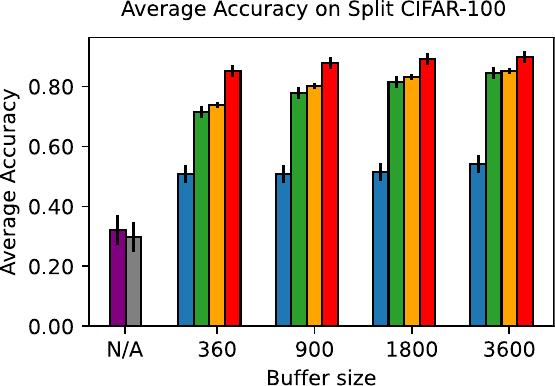} \\
         \vspace{0.2 cm}
         \caption{AA ($\uparrow$ better)}
     \end{subfigure}
     \hfill
     \begin{subfigure}[b]{0.31\textwidth}
         \centering
         \includegraphics[width=\textwidth]{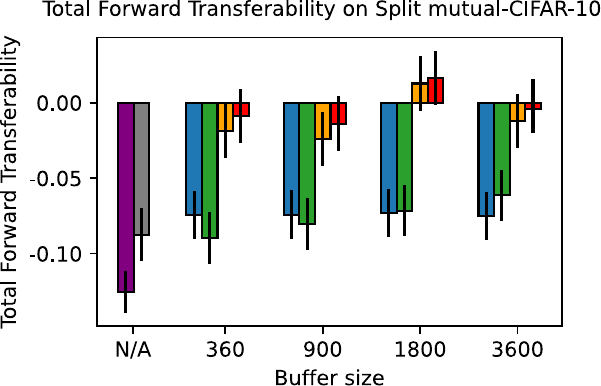} \\
         \vspace{0.3 cm}
         \includegraphics[width=\textwidth]{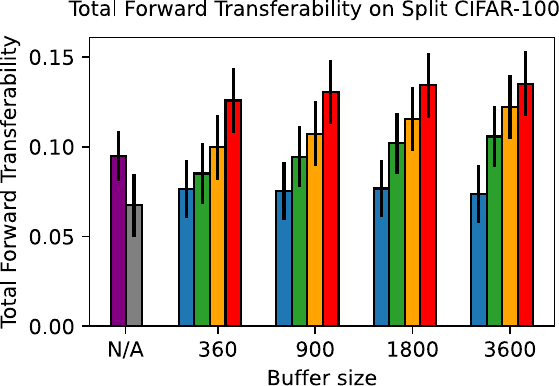} \\
         \vspace{0.2 cm}
         \caption{TFT ($\uparrow$ better)}
     \end{subfigure}
     \hfill
     \begin{subfigure}[b]{0.31\textwidth}
         \centering
         \includegraphics[width=\textwidth]{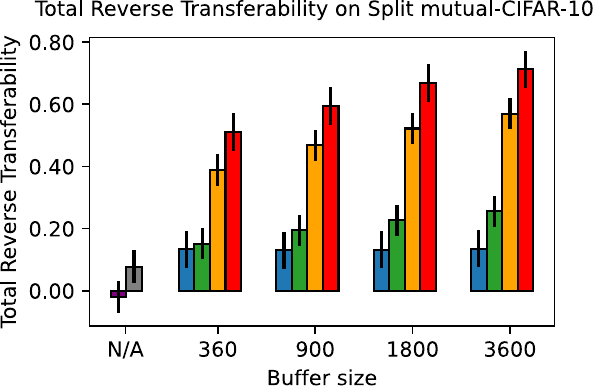} \\
         \vspace{0.3 cm}
         \includegraphics[width=\textwidth]{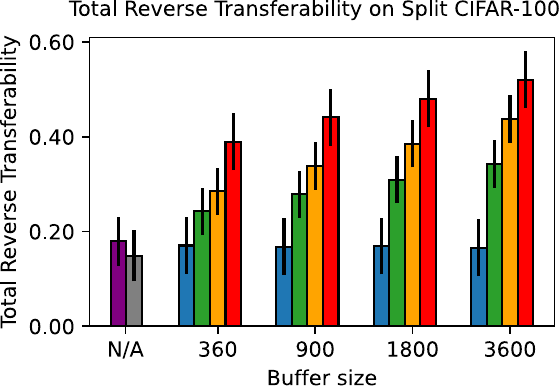} \\
         \vspace{0.2 cm}
         \caption{TRT ($\uparrow$ better)}
     \end{subfigure}
        \caption{The AA, TFT, and TRT of different CL algorithms on Split mutual-CIFAR-10 and Split CIFAR-100. Error bars are the standard deviations of the experiments.}
        \label{fig:score:no_buffer}
\end{figure*}

In the second experiment, we compare the AA of our method when changing the base transferability metric from LogME to LEEP~\citep{nguyen2020leep}, GBC~\citep{pandy2022transferability}, and TransRate~\citep{huang2022frustratingly}. Figure~\ref{fig:task_selection_transfer} gives the results of this experiment for both the one-batch and multiple-batch settings. From the figure, we can see that the average accuracies differ slightly when changing the transferability metric, with LogME achieving the best performance overall. Nevertheless, our method outperforms the random baseline for all choices of base transferability metrics.

\section{Conclusion}

We proposed two measures of sequence transferability for CL that can capture the total transferability in either forward or backward direction. Our experiments showed that these measures had close associations with CL accuracy. Based on these findings, we developed a heuristic task order selection method for CL and showed that it outperformed the random selection strategy while being robust to the sample size and the choice of its base transferability measure.

\appendix

\section{Additional Experiment Results for Regularization-based Methods}
\label{subsec:non-replay}

In Section \ref{sec:seqtrans-exp}, we empirically show the associations between AA and TFT/TRT for several common replay-based CL algorithms. In this appendix, we provide additional results for two regularization-based approaches, Learning without Forgetting (LwF) \citep{li2017learning} and Synaptic Intelligence (SI) \citep{zenke2017continual}. LwF uses the distillation loss as a regularizer to prevent losing information from previous tasks. SI, on the other hand, is motivated by the human brain's structure and determines the importance of each parameter using the entire learning trajectory.

We repeat the experiments in Section \ref{sec:seqtrans-exp} for these two methods on the Split mutual-CIFAR-10 and Split CIFAR-100 benchmarks. We use SGD with a batch size of 32 and a learning rate of 0.03 to train ResNet18 models \citep{he2016deep} for all methods. Each dataset undergoes model training for 50 epochs per task. The results of our experiments are reported in Figure~\ref{fig:score:no_buffer}. We can clearly observe that TFT and TRT have similar trends with AA for the two regularization-based methods. This observation is consistent with our claims in the main text.






\bibliographystyle{named}
\bibliography{ijcai25}

\end{document}